\documentclass[10pt,twocolumn,letterpaper]{article}

\usepackage[pagenumbers]{cvpr}

\usepackage[accsupp]{axessibility}
\usepackage{graphicx}
\usepackage{amsmath}
\usepackage{amssymb}
\usepackage{booktabs}[table]
\usepackage{makecell}

\usepackage[ruled,linesnumbered]{algorithm2e}
\usepackage{multirow}
\usepackage{balance}

\usepackage[pagebackref,breaklinks,colorlinks]{hyperref}

\usepackage[capitalize]{cleveref}
\crefname{section}{Sec.}{Secs.}
\Crefname{section}{Section}{Sections}
\Crefname{table}{Table}{Tables}
\crefname{table}{Tab.}{Tabs.}

\def\cvprPaperID{654}
\def\confName{CVPR}
\def\confYear{2023}

\begin{document}

\title{Uncovering the Missing Pattern: \\ Unified Framework Towards Trajectory Imputation and Prediction}

\author{Yi Xu\textsuperscript{1,2,\thanks{Work done during Yi's internship at Honda Research Institute, under Chiho Choi's supervision.}} \quad Armin Bazarjani\textsuperscript{2,3} \quad Hyung-gun Chi\textsuperscript{2,4} \quad Chiho Choi\textsuperscript{2,5} \quad Yun Fu\textsuperscript{1}\\
\textsuperscript{1}Northeastern University \quad
\textsuperscript{2}Honda Research Institute, USA \\
\textsuperscript{3}University of Southern California \quad
\textsuperscript{4}Purdue University \quad
\textsuperscript{5}Samsung Semiconductor US  \\
{\tt\small xu.yi@northeastern.edu, bazarjan@usc.edu, hgchi@purdue.edu} \\{\tt\small chiho1.choi@samsung.com, yunfu@ece.neu.edu}}
\maketitle  

\begin{abstract}
Trajectory prediction is a crucial undertaking in understanding entity movement or human behavior from observed sequences. However, current methods often assume that the observed sequences are complete while ignoring the potential for missing values caused by object occlusion, scope limitation, sensor failure, etc. This limitation inevitably hinders the accuracy of trajectory prediction. To address this issue, our paper presents a unified framework, the Graph-based Conditional Variational Recurrent Neural Network (GC-VRNN), which can perform trajectory imputation and prediction simultaneously. Specifically, we introduce a novel Multi-Space Graph Neural Network (MS-GNN) that can extract spatial features from incomplete observations and leverage missing patterns. Additionally, we employ a Conditional VRNN with a specifically designed Temporal Decay (TD) module to capture temporal dependencies and temporal missing patterns in incomplete trajectories. The inclusion of the TD module allows for valuable information to be conveyed through the temporal flow. We also curate and benchmark three practical datasets for the joint problem of trajectory imputation and prediction. Extensive experiments verify the exceptional performance of our proposed method. As far as we know, this is the first work to address the lack of benchmarks and techniques for trajectory imputation and prediction in a unified manner. 
\end{abstract}

\section{Introduction}\label{sec:intro}
\begin{figure}[t]
  \centering
   \includegraphics[width=0.98\linewidth]{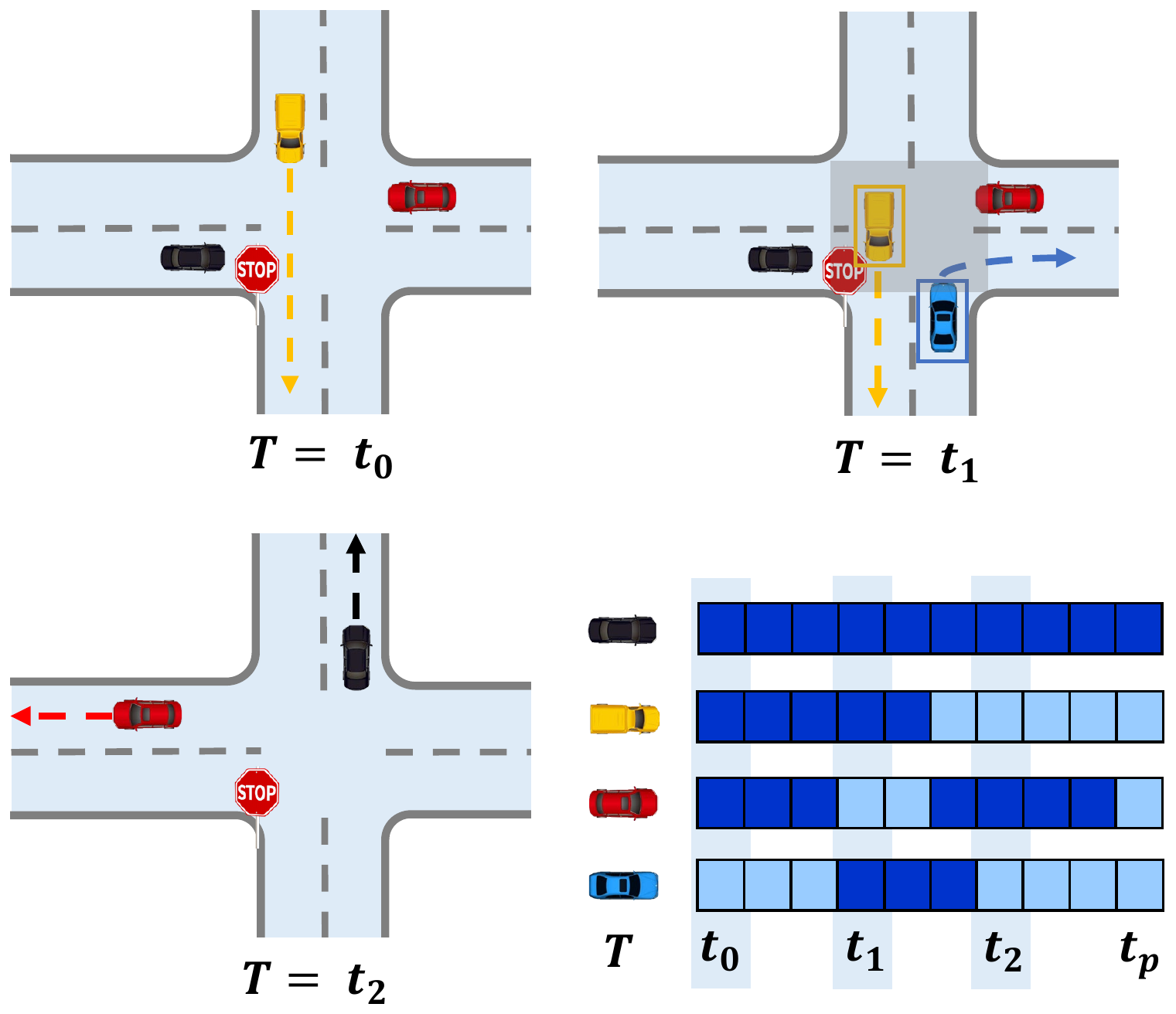}
   \vspace{-1mm}
   \caption{A typical example of incomplete observed trajectory. The black car (ego-vehicle) is waiting at the intersection at time step $t_{0}$, and the yellow car is moving. At time step $t_{1}$, the red car is occluded by the yellow car, and the blue car appears. The bottom right figure indicates the ``visibility'' of four cars, where dark means visible and light color means invisible.}
   \vspace{-3mm}
   \label{fig:introvehicle}
\end{figure}
\begin{figure*}[h]
  \centering
   \includegraphics[width=0.98\linewidth]{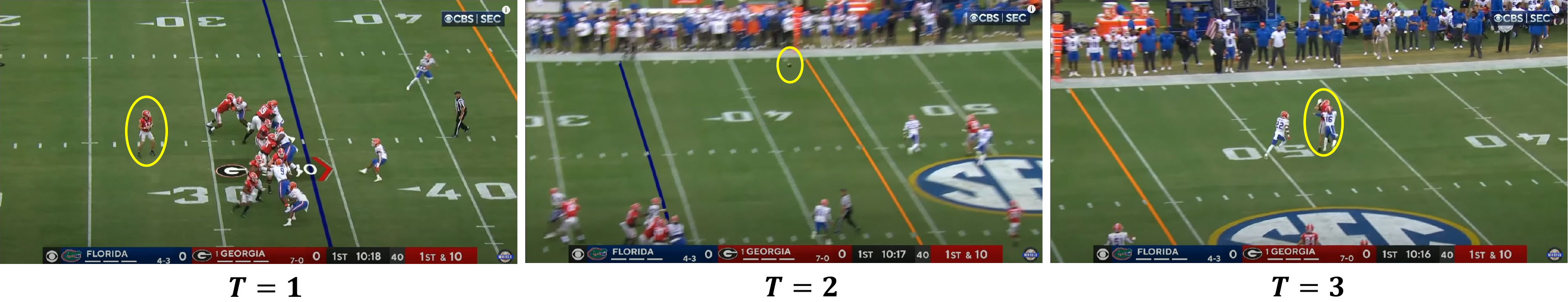}
   \vspace{-1mm}
   \caption{Three continuous frames of a ``throw-and-catch'' sequence in a live football match, where the ball is being circled. During this attacking play, the camera follows and zooms in on the ball. Only a subset of players is visible within the camera's field of view.}
   \vspace{-3mm}
   \label{fig:introfootball}
\end{figure*}

Modeling and predicting future trajectories play an indispensable role in various applications, i.e., autonomous driving~\cite{wang2022ltp, hazard2022importance, hu2022st}, motion capture~\cite{suwajanakorn2017synthesizing, taylor2017deep}, behavior understanding~\cite{girase2021loki, ma2021continual}, \textit{etc}. However, accurately predicting movement patterns is challenging due to their complex and subtle nature. Despite significant attention and numerous proposed solutions~\cite{xu2022remember, xu2022socialvae, tsao2022social, xu2022groupnet, li2022graph, bae2022learning} to the trajectory prediction problem, existing methods often assume that agent observations are entirely complete, which is too strong an assumption to satisfy in practice.

For example, in sports games such as football or soccer, not all players are always visible in the live view due to the limitation of the camera view. In addition, the live camera always tracks and moves along the ball, resulting in some players appearing and disappearing throughout the view, depending on their relative locations to the ball. ~\cref{fig:introfootball} illustrates this common concept. Similar situations also arise in autonomous driving where occlusion or sensor failure can cause missing data. As illustrated in~\cref{fig:introvehicle}, at time step $t_0$, there is no observation of the blue car. At time step $t_1$, the red car is occluded by the yellow car from the black car perspective, and the blue car appears at the intersection to turn right. Predicting future trajectories of entities under these circumstances will no doubt hinder the performance and negatively influence behavior understanding of moving agents or vehicle safety operations.

Although various recent works~\cite{yoon2018gain, fedus2018maskgan, luo2018multivariate, yi2019not, shukla2020multi, cini2021multivariate} have investigated the time series imputation problem, most are autoregressive models that impute current missing values from previous time steps, making them highly susceptible to compounding errors for long-range temporal modeling. Additionally, commonly used benchmarks~\cite{sezer2020financial, silva2012predicting, johnson2016mimic, zheng2014urban} do not contain interacting entities. Some recent works~\cite{liu2019naomi, zhan2018generating} have studied the imputation problem in a multi-agent scenario. Although these methods have achieved promising imputation performance, they fail to explore the relevance between the trajectory imputation and the prediction task. In fact, complete trajectory observation is essential for prediction, and accurate trajectory prediction can offer valuable information about the temporal correlation between past and future, ultimately aiding in the imputation process.

In this paper, we present a unified framework, Graph-based Conditional Variational Recurrent Neural Network (GC-VRNN), that simultaneously handles the trajectory imputation and prediction. Specifically, we introduce a novel Multi-Space Graph Neural Network (MS-GNN) to extract compact spatial features of incomplete observations. Meanwhile, we adopt a Conditional VRNN (C-VRNN) to model the temporal dependencies, where a Temporal Decay (TD) module is designed to learn the missing patterns of incomplete observations. The critical idea behind our method is to acquire knowledge of the spatio-temporal features of missing patterns, and then unite these two objectives through shared parameters. Sharing valuable information allows these two tasks to support and promote one another for better performance mutually. In addition, to support the joint evaluation of multi-agent trajectory imputation and prediction, we curate and benchmark three practical datasets from different domains, \textbf{\textit{Basketball-TIP}}, \textbf{\textit{Football-TIP}}, and \textbf{\textit{Vehicle-TIP}}, where the incomplete trajectories are generated via reasonable and practical strategies. The main contributions of our work can be summarized as follows:
\begin{itemize}
    \vspace{-1mm}
    \item We investigate the multi-agent trajectory imputation and prediction problem and develop a unified framework, GC-VRNN, for imputing missing observations and predicting future trajectories simultaneously.
    \vspace{-1mm}
    \item We propose a novel MS-GNN that can extract comprehensive spatial features of incomplete observations and adopt a C-VRNN with a specifically designed TD module for better learning temporal missing patterns, and valuable information is shared via temporal flow.
    \vspace{-1mm}
    \item We curate and benchmark three datasets for the multi-agent trajectory imputation and prediction problem. Strong baselines are set up for this joint problem.
    \vspace{-1mm}
    \item Thorough experiments verify the consistent and exceptional performance of our proposed method. 
\end{itemize}
\vspace{-1mm}

\begin{figure*}[ht]
  \centering
   \includegraphics[width=0.98\linewidth]{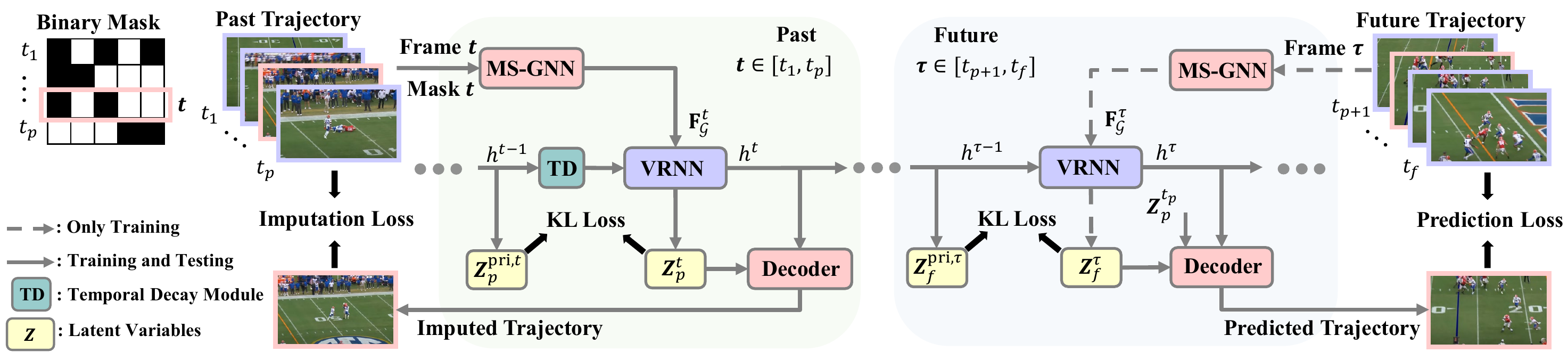}
   \vspace{-1mm}
   \caption{Overview of our GC-VRNN. The inputs are the past incomplete trajectory, the corresponding mask, and the future trajectory (only used for training). The outputs are imputed trajectory and predicted trajectory. Our model jointly handles the imputation and prediction problem, meanwhile, is trained in an end-to-end fashion.}
   \vspace{-3mm}
   \label{fig:overview}
\end{figure*}

\section{Related Work}
\subsection{Trajectory Prediction}
The objective of trajectory prediction is to predict the future positions of agents conditioned on their observations. A pioneering study, Social-LSTM~\cite{alahi2016social}, introduces a pooling layer that facilitates the sharing of human-human interaction features. Following this, some methods~\cite{vemula2018social, zhang2019sr, hu2020collaborative, xu2021tra2tra} have been proposed to extract comprehensive interaction features. Considering the uncertainty of human trajectory, some works use generative models such as Generative Adversarial Networks (GANs) ~\cite{gupta2018social, sadeghian2019sophie, li2019conditional, amirian2019social, kosaraju2019social} and Variational Autoencoders (VAEs) ~\cite{mangalam2020It, xu2022remember, ivanovic2019trajectron, salzmann2020trajectronplusplus, xu2022socialvae} to generate multiple trajectory predictions. Recently, Transformer structure~\cite{vaswani2017attention} is applied in this task~\cite{yuan2021agentformer, tsao2022social, giuliari2020transformer, yu2020spatio} to model the spatio-temporal relations via an attention mechanism. Moreover, various viewpoints have emerged towards more practical applications, i.e., goal-driven idea~\cite{tran2021goal, zhao2021you, mangalam2021goals, chiara2022goal}, long-tail situation~\cite{makansi2021exposing}, interpretability~\cite{kothari2021interpretable}, robustness~\cite{xu2021robust, weng2022whose, zhang2022adversarial, cao2022advdo}, counterfactual analysis~\cite{chen2021human}, planning-driven~\cite{chen2022scept}, generalization ability to new environment~\cite{xu2022adaptive, ivanovic2022expanding, bahari2022vehicle}, and knowledge distillation~\cite{monti2022many}.

Typically, in graph-based models~\cite{kosaraju2019social, sun2020recursive, mohamed2020social, shi2021sgcn, xu2022groupnet, li2022graph, bae2022learning}, each individual is considered as a single node, while the connections between them are depicted as edges. Graph Convolutional Layers (GCLs) and a Message Passing (MP) mechanism are utilized to extract spatio-temporal characteristics. However, these methods presume that observations are complete, which can be difficult to meet in real-world situations. Furthermore, these graph-based models are unable to identify missing patterns, whereas our proposed approach can reveal incomplete spatio-temporal patterns.

\subsection{Trajectory Imputation}
Some statistical imputation techniques substitute missing values with the mean or median value~\cite{acuna2004treatment}. Other alternatives, such as linear fit~\cite{ansleyestimation}, k-nearest neighbours~\cite{troyanskaya2001missing, beretta2016nearest}, and expectation-maximization (EM) algorithm~\cite{ghahramani1993supervised, nelwamondo2007missing}, are also adopted. One of the biggest limitations of such methods is using rigid priors, which hinders the generalization ability. A more flexible framework is utilized with generative methods to learn the missing pattern. For instance, some deep autoregressive methods based on RNNs~\cite{lipton2016directly, che2018recurrent, yoon2018estimating, cao2018brits} are proposed to impute the sequential data. Some other methods~\cite{yoon2018gain, luo2018multivariate, qi2020imitative, fortuin2020gp, miao2021generative, tashiro2021csdi} have been proposed to leverage GANs or VAEs to generate reconstructed sequences. 

Few works explore the trajectory imputation problem in the multi-agent scenario. Notably, NAOMI~\cite{liu2019naomi} presents a non-autoregressive imputation method that exploits the multi-resolution structure of sequential data for imputation. GMAT~\cite{zhan2018generating} designs a hierarchical model to produce weak macro-intent labels for sequence generation. However, these two methods only focus on the trajectory imputation task and fail to investigate the prediction task. In work INAM~\cite{qi2020imitative}, an imitation learning paradigm is proposed to handle the imputation and prediction in an asynchronous mode. While our model handles these two tasks simultaneously and is trained in an end-to-end fashion. Furthermore, method INAM is solely assessed on a single multi-agent dataset, with missing instances being generated at random. We argue that this arbitrary masking technique is not practical in real-world scenarios. Conversely, our work has been validated as effective across various multi-agent domains. Most importantly, we also leverage missing patterns in the realm of spatio-temporal features.

\section{Problem Definition}
Consider an observed set of $N$ agents $\Omega=\{1, 2,..., N\}$ over time step $t_{1}$ to $t_{p}$. Let $X_{i}^{\leq t_{p}}=\{\boldsymbol{x}_{i}^{1},..., \boldsymbol{x}_{i}^{t},..., \boldsymbol{x}_{i}^{t_{p}}\}$ denote the observed trajectory of agent $i$, where $\boldsymbol{x}_{i}^{t} \in \mathbb{R}^{2}$ represents the 2D coordinates of agent $i$ at time step $t$. The observed trajectory set is thus defined as $X_{\Omega}^{\leq t_{p}}=\{X_{i}^{\leq t_{p}}|\forall i \in \Omega\}$. Because some observations for any subset of agents could be missing at any time due to occlusion, sensor failure, \textit{etc}. The missing locations are represented by a masking matrix $M_{i}^{\leq t_{p}}=\{\boldsymbol{m}_{i}^{1},..., \boldsymbol{m}_{i}^{t},..., \boldsymbol{m}_{i}^{t_{p}}\}$ valued in $\{0, 1\}$. The variable $\boldsymbol{m}_{i}^{t}$ is assigned a value of $1$ if the observation is available at time step $t$ and $0$ otherwise. 

The goal of the joint problem of multi-agent trajectory imputation and prediction is to impute missing values of all agents observations from time step $t_{1}$ to $t_{p}$, and also predict their future trajectory from time step $t_{p+1}$ to $t_{f}$ conditioned on their incomplete observations. More formally, that is to learn a model $f(\cdot)$ with parameter $W^{*}$ that outputs $\hat{X}_{\Omega}^{\leq t_{p}}$ and $\hat{Y}_{\Omega}^{t_{p+1} \leq t \leq t_{f}}$, where $\hat{X}_{\Omega}^{\leq t_{p}}$ refers to the imputed trajectory and $\hat{Y}_{\Omega}^{t_{p+1} \leq t \leq t_{f}}$ refers to the predicted trajectory.

\section{Proposed Method}
\cref{fig:overview} illustrates our proposed method at a high level. The fundamental idea is to address the task of multi-agent trajectory imputation and prediction in a unified framework while promoting information exchange via temporal flow. 
\subsection{MS-GNN}
Applying a graph-based approach is a natural decision to model the spatial correlations of multiple agents. However, as we are addressing the unique challenge of incomplete observations. In order to extract spatial features from incomplete observations of multiple agents, we develop a novel approach called Multi-Space Graph Neural Network (MS-GNN) that enhances the capability of Graph Convolutional Layers (GCL).

\begin{figure}[t]
  \centering
   \includegraphics[width=0.98\linewidth]{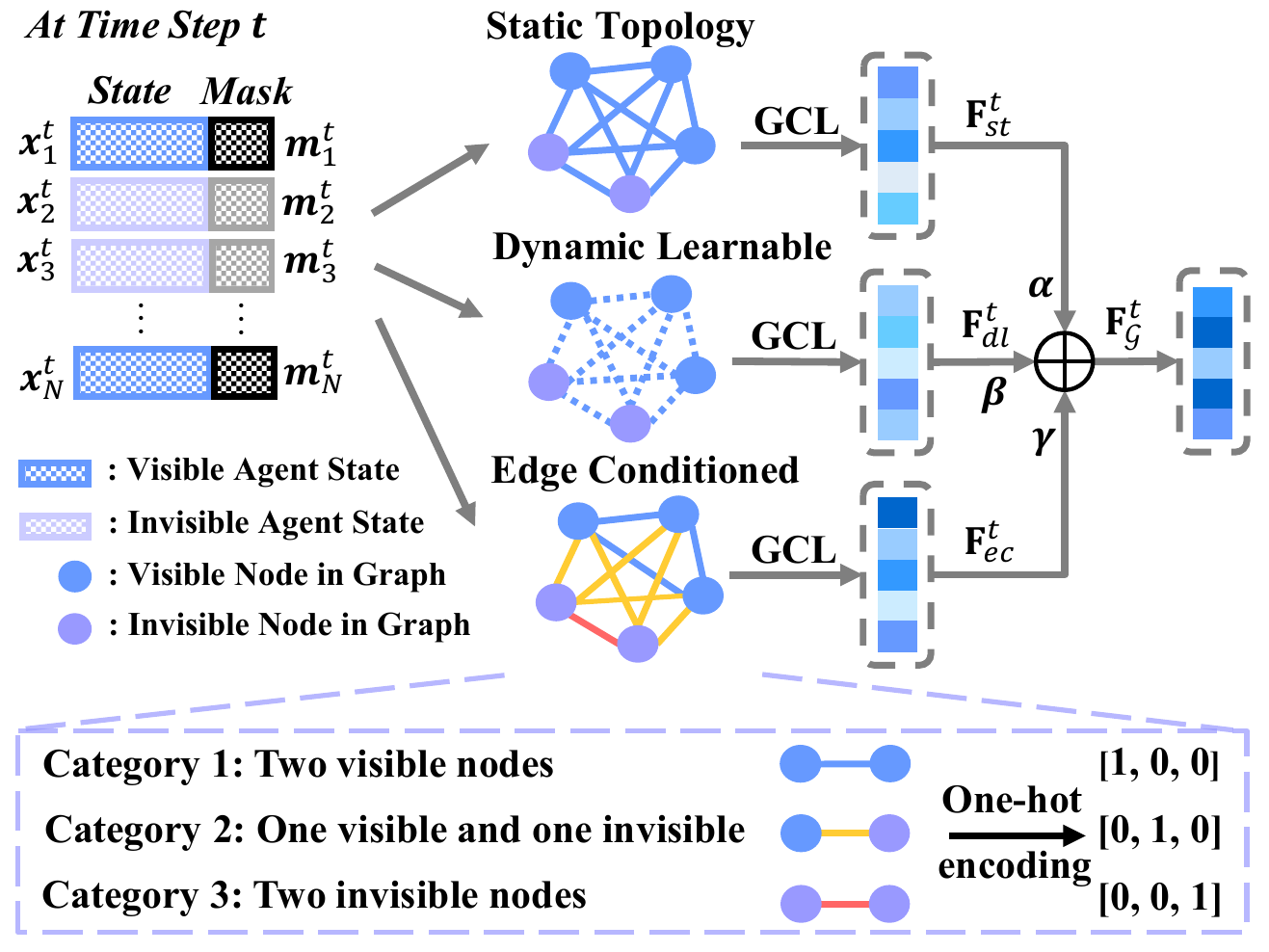}
   \vspace{-1mm}
   \caption{Schematic diagram of three different agent-wise graph convolutional layers at time step $t$. The output features of three GCLs are finally integrated to $\mathbf{F}_{\mathcal{G}}^{t}$ via weighted sum.}
   \vspace{-3mm}
   \label{fig:msgnn}
\end{figure}

\textbf{Graph Construction.}
Each agent is considered as a single node in the graph, and the graph at time step $t$ is defined as $\mathcal{G}^t=(\mathcal{V}^{t}, \mathcal{E}^{t})$, where $\mathcal{V}^{t}=\{v^{t}_{i}|i \in\Omega\}$ denotes the vertex set of agents, $\mathcal{E}^{t}=\{e^{t}_{i,j}|i,j\in\Omega\}$ denotes the edge set captured by an adjacency matrix $\mathbf{A}^{t}=\{a^{t}_{i,j}|i,j\in\Omega\}$. The graph feature representation is defined as $\mathbf{F}^{t}=\{\boldsymbol{f}^{t}_{i}\in\mathbb{R}^{D}|i\in\Omega\}$, where $\boldsymbol{f}^{t}_{i}$ is the feature vector of node $i$ at time step $t$, $D$ denotes the dimension of node feature vector.

\textbf{Graph Input.}
The inputs of MS-GNN are the incomplete observed trajectory, the corresponding binary mask that indicates the missing status, and the future trajectory. Note that the future trajectory is only used in the training phase. For instance, consider agent $i$ at time step $t$, the node feature is initialized by projecting inputs to high-dimensional feature vectors, which are defined as follows:
\begin{equation}
    \boldsymbol{f}_{i}^{t} = \left\{
    \begin{aligned}
    &\varphi_{p} \left((\boldsymbol{x}_{i}^{t}\odot \boldsymbol{m}_{i}^{t})\oplus \boldsymbol{m}_{i}^{t}; \mathbf{W}_{p}\right)\quad t\in \left[t_{1}, t_{p}\right]\\
    &\varphi_{f} \left(\boldsymbol{y}_{i}^{t}; \mathbf{W}_{f}\right)\quad t\in \left[t_{p+1}, t_{f}\right]
    \end{aligned}
    \right. ,
    \label{eq:input}
\end{equation}
where $\varphi_{p}(\cdot)$ and $\varphi_{f}(\cdot)$ are different projection functions with weights $\mathbf{W}_{p}\in\mathbb{R}^{3\times D}$ and $\mathbf{W}_{f}\in\mathbb{R}^{2\times D}$, respectively. In our implementation, we achieve this using MLPs. $\odot$ denotes the element-wise multiplication, and $\oplus$ denotes the concatenation operation. $\boldsymbol{y}_{i}^{t}$ represents the future location of agent $i$, which is only used in the training phase.

In MS-GNN, we define three different GCLs to extract spatial features from different feature spaces at each time step. Each GCL is designed to extract primary features for intuitive purposes, and meanwhile emphasize the spatial correlations of missing patterns in observations. To avoid confusion and for simplicity, we omit the \textbf{superscript} $t$ of node feature $\boldsymbol{f}_{i}^{t}$ as $\boldsymbol{f}_{i}$, and graph feature $\mathbf{F}^{t}$ as $\mathbf{F}$.

\textbf{Static Topology GCL.} In the vanilla
GCL~\cite{kipf2016semi}, the adjacency matrix only indicates the connectivity of node pairs, where $\mathbf{A}_{i,j}=1$ if an edge directs from node $i$ to $j$ and $0$ otherwise. While in our case, some of the agents (nodes) are missing in the observed trajectory. Therefore, we define a different adjacency matrix $\mathbf{A}_{st}$ not only to indicate the connectivity but also the visibility in static topology GCL. The identity matrix $\mathbf{I}_{st}$ is adjusted accordingly with a constraint for adding the self-loop. Constraints are defined as follows:

\noindent \textit{\textbf{Constraint 1:} $\mathbf{A}^{t}_{i,j}=\mathbf{A}^{t}_{j,i}=1$ if node $i$ and $j$ are both visible at time step $t$. Otherwise, $\mathbf{A}^{t}_{i,j}=\mathbf{A}^{t}_{j,i}=0$.}

\noindent \textit{\textbf{Constraint 2:} $\mathbf{I}^{t}_{i,i}=1$ if node $i$ is visible at time step $t$. Otherwise, $\mathbf{I}^{t}_{i,i}=0$.}

Similar to~\cite{kipf2016semi}, the propagation rule of graph feature $\mathbf{F}^{(l)}_{st}$ of the $l$-th static topology layer is defined as follows:
\begin{equation}
    \mathbf{F}^{(l+1)}_{st} = \sigma \left(\mathbf{\hat{A}}_{st}\mathbf{F}^{(l)}_{st}\mathbf{W}^{(l)}_{st}\right),
    \label{eq:stgcl}
\end{equation}
where normalized adjacency matrix $\mathbf{\hat{A}}_{st}=\mathbf{\widetilde{D}}^{-1/2}(\mathbf{A}_{st}+\mathbf{I}_{st})\mathbf{\widetilde{D}}^{1/2}$, $\mathbf{\widetilde{D}}$ is the diagonal degree of $\mathbf{A}_{st}+\mathbf{I}_{st}$, $\mathbf{W}^{(l)}_{st}\in \mathbb{R}^{D^{(l)} \times D^{(l+1)}}$ are learnable parameters of the $l$-th static topology layer, and $\sigma(\cdot)$ denotes the ReLU activation function. This static topology GCL models the connectivity and visibility features of agents in a fixed way.

\textbf{Dynamic Learnable GCL.} In contrast to the $\mathbf{A}_{st}$ in topology GCL with fixed values ($0$ or $1$), inspired by~\cite{shi2019two}, we define a simple, learnable, and unconstrained $\mathbf{A}_{dl}$ to dynamically learn the strength of relations between nodes, and to improve the flexibility of GCL. The matrix $\mathbf{A}_{dl}$ is initialized with random values and is trained to modify the edges by either strengthening, weakening, adding, or removing them. Similar to~\cref{eq:stgcl}, the propagation rule of graph feature $\mathbf{F}^{(l)}_{dl}$ is defined as follows: 
\begin{equation}
    \mathbf{F}^{(l+1)}_{dl} = \sigma \left(\mathbf{A}_{dl}\mathbf{F}^{(l)}_{dl}\mathbf{W}^{(l)}_{dl}\right),
    \label{eq:dl}
\end{equation}
where $\mathbf{W}^{(l)}_{dl}\in \mathbb{R}^{D^{(l)} \times D^{(l+1)}}$ are learnable parameters of the $l$-th dynamic learnable layer. Since all the elements in $\mathbf{A}_{dl}$ are learnable with no constraint, the $\mathbf{A}_{dl}$ will be asymmetric that allows each edge to select the best suitable relation strength to update its corresponding node features. Intuitively, compared to the static topology GCL, the relations among agents (nodes) are better captured by this GCL.

\textbf{Edge Conditioned GCL.} The aforementioned two GCLs focus on learning spatial relations among nodes and the different strengths of such relations by two different definitions of the adjacency matrix. While in our case, one challenge is that node features of some agents are missing in the incomplete observations. In order to better understand the spatial missing patterns, we leverage an edge conditioned GCL, where we assign a label to each edge based on its category and integrate such category information in graph propagation. As shown in~\cref{fig:msgnn}, three types of edges exist, which are determined by the visibility of the corresponding node pair. We first encode three categories into one-hot vectors $\ \vartheta_{i,j}\in\mathbb{R}^{3}$, and then define a mapping network $\varphi_{ec}(\cdot)$ to output the edge-specific weight matrix $\Theta_{i,j}\in \mathbb{R}^{D_{\mathcal{G}} \times D}$ for updating the node features. The updating rule in edge conditioned GCL is defined as:
\begin{equation}
\begin{aligned}
    \boldsymbol{f}_{ec;i} &= \frac{1}{|\mathcal{V}(i)|}\sum_{j\in\mathcal{V}(i)}\varphi_{ec}(\vartheta_{i,j}; \mathbf{W}_{ec})\boldsymbol{f}_{j} + \mathbf{b}_{ec} \\
    &= \frac{1}{|\mathcal{V}(i)|}\sum_{j\in\mathcal{V}(i)}\Theta_{i,j}\boldsymbol{f}_{j} + \mathbf{b}_{ec}
\end{aligned},
\label{eq:edge}
\end{equation}
where $\boldsymbol{f}_{ec;i}$ denotes the feature vector for node $i$ in $\mathbf{F}_{ec}$, while $\mathbf{W}_{ec}$ represents the learnable parameters of the $l$-th edge conditioned layer, $\mathbf{b}_{ec}$ is the learnable bias, and $\mathcal{V}(i)$ denotes the neighboring nodes set of node $i$. In our implementation, we utilize a Conv2D block consisting of two Conv2D layers and one average pooling layer. Note that this GCL is only employed for the observations.

\textbf{Graph Feature Fusion.} Upon obtaining the last layer graph feature representations, namely $\mathbf{F}_{st}$, $\mathbf{F}_{dl}$, and $\mathbf{F}_{ec}$, we integrate them into a final graph representation denoted as $\mathbf{F}_{\mathcal{G}} \in \mathbb{R}^{N \times D_{\mathcal{G}}}$ through the following equation. This is achieved by setting the feature dimension of the final layer of each GCL as $D_{\mathcal{G}}$.
\begin{equation}
    \mathbf{F}_{\mathcal{G}}=\alpha \mathbf{F}_{st}+\beta \mathbf{F}_{dl}+\gamma \mathbf{F}_{ec},
    \label{eq:gff}
\end{equation}
where $\alpha$, $\beta$, and $\gamma$ are three learnable parameters for feature fusion with the same size $\mathbb{R}^{D_{\mathcal{G}}}$.

\subsection{C-VRNN with TD}
In our work, we leverage a C-VRNN for modeling both past and future trajectory temporal dependencies. In the imputation stream, we introduce a Temporal Decay (TD) module that is dependent on the time interval between the preceding observation and the current time step. Furthermore, we recurrently update the priors of trajectory and the latent variables of imputation and prediction streams via a parameter-shared temporal flow. Valuable information is promoted to exchange implicitly with one another.

The vanilla VRNN~\cite{chung2015recurrent} can be considered as a basic VAE conditioned on the hidden states of an RNN and it is trained by maximizing the Sequential ELBO as follows:
\begin{equation}
\begin{split}
    \mathbb{E}_{q_{\phi}(\boldsymbol{z}^{\leq T}|\boldsymbol{x}^{\leq T})}
    \Bigg[ &\sum^{T}_{t=1}\log p_{\theta}(\boldsymbol{x}^{t}|\boldsymbol{z}^{\leq t}, \boldsymbol{x}^{<t})\\
    -&\text{KL}\left(q_{\phi}(\boldsymbol{z}^{t}|\boldsymbol{x}^{\leq t}, \boldsymbol{z}^{<t})||
    p_{\theta}(\boldsymbol{z}^{t}|\boldsymbol{x}^{<t}, \boldsymbol{z}^{<t})\right)\Bigg].
\end{split}
\label{eq:elbo}
\end{equation}

\textbf{Prior.}
The distribution for the prior on the latent variables $\boldsymbol{z}^{t}$ follows the following format at each time step:
\begin{equation}
    \boldsymbol{z}^{t} \sim \mathcal{N}( \boldsymbol{\mu}^{\text{pri},t},{\boldsymbol{\sigma}^{\text{pri},t}}^2),
\end{equation}
where the distribution parameters $\boldsymbol{\mu}^{\text{pri},t}$ and ${\boldsymbol{\sigma}^{\text{pri},t}}^{2}$ are conditioned on the hidden states $h^{t-1}$ of RNN as follows:
\begin{equation}
    [\boldsymbol{\mu}^{\text{pri},t}, {\boldsymbol{\sigma}^{\text{pri},t}}^{2}] = \varphi^{\text{pri}}\left(h^{t-1}; \mathbf{W}^{\text{pri}}\right),
\end{equation}
where $\varphi^{\text{pri}}(\cdot)$ is a mapping function that maps hidden state to a prior distribution with weights $\mathbf{W}^{\text{pri}}$.

\textbf{Generation.} 
At time step $t$, the generation process aims to decode imputed trajectory or future prediction from latent variables. Similarly, we assume that the location (2D coordinates) of agents follows a bi-variate Gaussian distribution as $\boldsymbol{x}^{t}\sim\mathcal{N}(\boldsymbol{\mu}^{t}, \boldsymbol{\sigma}^{t}, \boldsymbol{\rho}^{t})$, where $\boldsymbol{\mu}^{t}$ is the mean, $\boldsymbol{\sigma}^{t}$ is the standard deviation, and $\boldsymbol{\rho}^{t}$ is the correlation coefficient. 

For imputation, the generating distribution is conditioned on $\boldsymbol{z}^{t}$ and the previous hidden state $h^{t-1}$ such that: 
\begin{equation}
\begin{split}
    [\hat{\boldsymbol{\mu}}^{t}, \hat{\boldsymbol{\sigma}}^{t}, \hat{\boldsymbol{\rho}}^{t}] =   \varphi^{\text{dec}}_{p}\left(
    \varphi^{\boldsymbol{z}}_{p}\left(
    \boldsymbol{z}^{t}\right) \oplus h^{t-1};  \mathbf{W}^{\text{dec}}_{p}\right) \\
    t\in \left[t_{1}, t_{p}\right]
\end{split}
.
\end{equation}

Differently, apart from $\boldsymbol{z}^{t}$ and the previous hidden state $h^{t-1}$, the predicting distribution is also conditioned on the latent variables $\boldsymbol{z}^{t_{p}}$ of the last observed time step such that:
\begin{equation}
\begin{split}
    [\hat{\boldsymbol{\mu}}^{t}, \hat{\boldsymbol{\sigma}}^{t}, \hat{\boldsymbol{\rho}}^{t}]=
    \varphi^{\text{dec}}_{f}\left(
    \varphi^{\boldsymbol{z}}_{f}\left(
    \boldsymbol{z}^{t}\oplus \boldsymbol{z}^{t_{p}}\right)\oplus h^{t-1}; \mathbf{W}^{\text{dec}}_{f}\right) \\ 
    t\in \left[t_{p+1}, t_{f}\right]
\end{split}
,
\end{equation}
where $\varphi^{\text{dec}}_{\cdot}(\cdot)$ is a decoding function with weights $\mathbf{W}^{\text{dec}}_{\cdot}$, and $\varphi^{\boldsymbol{z}}_{\cdot}(\cdot)$ is feature extractor of $\boldsymbol{z}^{t}$. In the prediction decoder, we first concatenate latent variable $\boldsymbol{z}^{t}$ to $\boldsymbol{z}^{t_{p}}$, and then extract the joint features. Intuitively, the feature information encoded by the imputation stream is also considered when decoding the predicting distribution.

\textbf{Temporal Decay.} 
In order to extract temporal features of missing patterns in observations, here we first introduce a temporal lag $\delta^{t}_{i}$ that indicates the relative distance between the last observable time step and the current time step $t$ of agent $i$. The temporal lag is calculated as follows:
\begin{equation}
    \delta^{t}_{i} = \left\{
    \begin{aligned}
        &t-(t-1)+\delta^{t-1}_{i} &\text{if $t>1$ and $\boldsymbol{m}^{t}_{i}=0$}\\
        &t-(t-1) &\text{if $t>1$ and $\boldsymbol{m}^{t}_{i}=1$} \\
        &0 &\text{if $t=1$}
    \end{aligned}
    \right. .
\end{equation}

Concatenate temporal lags $\delta^{t}_{i}$ of all the agents at time step $t$, we can obtain the temporal lag vector $\boldsymbol{\delta}^{t}$. Then, the temporal decay vector $\boldsymbol{\Delta}^{t}$ is calculated as follows:
\begin{equation}
    \boldsymbol{\Delta}^{t}= 1/\exp\left(\mbox{max}(0, \mathbf{W_{\delta}}\boldsymbol{\delta}^{t}+\mathbf{b_{\delta}})\right),
\end{equation}
where $\mathbf{W_{\delta}}$ and $\mathbf{b_{\delta}}$ are learnable parameters and bias. The insights behind this design lie in several points. In sequential modeling, if a variable has been missing for a while, its influence from the input will gradually decrease over time. Since the temporal lag $\boldsymbol{\delta}^{t}$ represents the distance from the last observation to the current time step, the temporal lag and temporal decay should be negatively correlated. Therefore, we chose a negative exponential function to ensure that the temporal decay decreases monotonically within a reasonable range of $0$ and $1$. Note that the decay vector is only calculated and applied for the past incomplete trajectory.

\textbf{Recurrence.} 
To capture more complex patterns from missing data, simply relying on temporal decay vectors may not be sufficient. Therefore, we propose to enhance the information obtained from the temporal decay vectors by element-wise multiplying them with the hidden states during the recurrence updating process as follows:
\begin{equation}
    {h^{t-1}}^{'} = \boldsymbol{\Delta}^{t}\odot h^{t-1},
\end{equation}
Intuitively, this operation can decay the extracted features rather than temporal decayed values. Finally, for the imputation stream, the RNN is updated as follows:
\begin{equation}
\begin{split}
 h^{t} =\text{RNN}\left(\left(
    \mathbf{F}^{t}_{\mathcal{G}}\oplus 
     \varphi^{\boldsymbol{z}}_{p}\left(
    \boldsymbol{z}^{t}\right)
    \right), {h^{t-1}}^{'}\right) \\
    t\in \left[t_{1}, t_{p}\right]
\end{split}
,
\end{equation}
while for the prediction stream, the RNN is updated as:
\begin{equation}
\begin{split}
 h^{t} =\text{RNN}\left(\left(
    \mathbf{F}^{t}_{\mathcal{G}}\oplus 
     \varphi^{\boldsymbol{z}}_{f}\left(
    \boldsymbol{z}^{t} \oplus \boldsymbol{z}^{t_{p}}\right)
    \right), {h^{t-1}}\right)\\
    t\in \left[t_{p+1}, t_{f}\right]
\end{split}
,
\end{equation}
where we also concatenate latent variable $\boldsymbol{z}^{t}$ to $\boldsymbol{z}^{t_{p}}$ when updating the hidden states in the prediction stream.

\textbf{Inference.} 
At each time step, the approximate posterior distribution of latent variables follows the distribution as:
\begin{equation}
\begin{aligned}
    &\boldsymbol{z}^{t}|\boldsymbol{x}^{t} \sim \mathcal{N}( \boldsymbol{\mu}^{\text{enc},t}, {\boldsymbol{\sigma}^{\text{enc},t}}^{2})
    \quad t\in \left[t_{1}, t_{p}\right] \\
    &\boldsymbol{z}^{t}|\boldsymbol{y}^{t} \sim \mathcal{N}( \boldsymbol{\mu}^{\text{enc},t}, {\boldsymbol{\sigma}^{\text{enc},t}}^{2})
    \quad t\in \left[t_{p+1}, t_{f}\right]
\end{aligned}
.
\end{equation}

The approximate posterior distribution is conditioned on graph representation and hidden states of RNN as follows:
\begin{equation}
    [\boldsymbol{\mu}^{\text{enc},t}, {\boldsymbol{\sigma}^{\text{enc},t}}^{2}] = \varphi^{\text{enc}}\left(\left(\mathbf{F}^{t}_{\mathcal{G}} \oplus h^{t-1}\right); \mathbf{W}^{\text{enc}}\right),
\end{equation}
where $\varphi^{\text{enc}}(\cdot)$ is the encoding function with weights $\mathbf{W}^{\text{enc}}$. 

\textbf{Loss Function.}
The loss function formed by~\cref{eq:elbo} of our proposed model consists of two parts: $\mathcal{L}_{\text{imp}}$ for imputation and $\mathcal{L}_{\text{pre}}$ for prediction, which are defined as follows:

\begin{equation}
\begin{split}
    \mathcal{L}_{\text{imp}} = - & \sum_{t=t_1}^{t_p}\log (\mathbb{P}(
    \boldsymbol{x}^{t}
    |\hat{\boldsymbol{\mu}}^{t}, \hat{\boldsymbol{\sigma}}^{t}, \hat{\boldsymbol{\rho}}^{t})) \\
    + & \lambda_{1} \text{KL}(
    \mathcal{N}(\boldsymbol{\mu}^{\text{enc},t}, {\boldsymbol{\sigma}^{\text{enc},t}}^{2})||
    \mathcal{N}(\boldsymbol{\mu}^{\text{pri},t}, {\boldsymbol{\sigma}^{\text{pri},t}}^{2}))  
\end{split}
,
\end{equation}
\begin{equation}
\begin{split}
    \mathcal{L}_{\text{pre}} = - & \sum_{t=t_{p+1}}^{t_f} \log (\mathbb{P}(
    \boldsymbol{y}^{t}
    |\hat{\boldsymbol{\mu}}^{t}, \hat{\boldsymbol{\sigma}}^{t}, \hat{\boldsymbol{\rho}}^{t})) \\
    + & \lambda_{2} \text{KL}(
    \mathcal{N}(\boldsymbol{\mu}^{\text{enc},t}, {\boldsymbol{\sigma}^{\text{enc},t}}^{2})||
    \mathcal{N}(\boldsymbol{\mu}^{\text{pri},t}, {\boldsymbol{\sigma}^{\text{pri},t}}^{2}))  
\end{split}
.
\end{equation}

The overall loss is defined in the following manner:
\begin{equation}
    \mathcal{L} = \mathcal{L}_{\text{imp}} + \lambda_{3} \mathcal{L}_{\text{pre}},
\end{equation}
where $\{\lambda_{1}, \lambda_{2}, \lambda_{3}\}$ are weighting factors, and we set them as $1$ in our model. Note that the loss is calculated over all agents in each trajectory, and our model is trained end-to-end for the joint problem of imputation and prediction.

\section{Experiments}\label{sec:exp}
\begin{table*}[t]
	\centering
    \scalebox{0.85}{
    \begin{tabular}{lc|cc|cc|cc|cc|cc|cc}
    \toprule
    \multicolumn{1}{c}{\multirow{2}{*}{Datasets}} & \multicolumn{1}{c}{\multirow{2}{*}{Methods}} & \multicolumn{2}{c}{$r=3$ ft.} &  \multicolumn{2}{c}{$r=5$ ft.} &  \multicolumn{2}{c}{$r=7$ ft.} & \multicolumn{2}{c}{$\theta=10^{\circ}$} & \multicolumn{2}{c}{$\theta=20^{\circ}$} & \multicolumn{2}{c}{$\theta=30^{\circ}$} 
    \\
     & \multicolumn{1}{c}{}  & I-$L_{2}$ & \multicolumn{1}{c}{P-$L_{2}$} & I-$L_{2}$ & \multicolumn{1}{c}{P-$L_{2}$} & I-$L_{2}$ & \multicolumn{1}{c}{P-$L_{2}$} & I-$L_{2}$ & \multicolumn{1}{c}{P-$L_{2}$} & I-$L_{2}$ & \multicolumn{1}{c}{P-$L_{2}$} & I-$L_{2}$ & \multicolumn{1}{c}{P-$L_{2}$}
     \\
    \midrule
    \multirow{9}{*}{\makecell[c]{\textbf{Basketball-TIP}\\ (In Feet)}}
    & Mean 
    & 9.07 & $-$ & 9.53 & $-$ & 9.51 & $-$  
    & 8.83 & $-$ & 8.64 & $-$ & 8.47 & $-$ 
    \\ 
    & Median 
    & 9.32 & $-$ & 9.82 & $-$ & 9.81 & $-$  
    & 9.16 & $-$ & 8.96 & $-$ & 8.75 & $-$  
    \\   
    & GMAT~\cite{zhan2018generating}
    & 7.36 & $-$ & 6.89 & $-$ & 6.73 & $-$   
    & 6.42 & $-$ & 5.99 & $-$ & 6.01 & $-$ 	
    \\  
    & NAOMI~\cite{liu2019naomi}
    & 7.68 & $-$ & 7.08 & $-$ & 7.04 & $-$  
    & 6.33 & $-$ & 6.11 & $-$ & 5.91 & $-$ 	
    \\
    & Linear Fit
    & 14.90 & 21.14 & 14.06 & 20.36 & 13.58 & 18.94
    & 12.78 & 21.01 & 11.47 & 16.38 & 11.26 & 14.40
    \\  
    & Vanilla LSTM~\cite{hochreiter1997long}
    & 7.33 & 20.07 & 6.73 & 14.91 & 6.51 & 10.07
    & 6.28 & 9.34 & 6.01 & 7.52 & 5.67 & 6.10
    \\
    & Vanilla VRNN~\cite{chung2015recurrent}
    & 7.43 & 12.26 & 6.90 & 11.38 & 6.68 & 10.07
    & 6.38 & 8.49 & 6.09 & 7.47 & 5.92 & 7.36
    \\
    & INAM~\cite{qi2020imitative}
    & 7.35 & 8.93 & 6.93 & 8.24 & 6.80 & 7.68
    & 6.50 & 7.32 & 6.13 & 7.10 & 5.92 & 6.96	
    \\   
    \cmidrule(lr){2-14}
    & \textbf{GC-VRNN} (\textbf{Ours}) 
    & \textbf{7.03} & \textbf{7.50} & \textbf{6.41} & \textbf{6.80} & \textbf{6.24} & \textbf{5.93} 
    & \textbf{5.86} & \textbf{6.29} & 
    \textbf{5.56} & \textbf{4.74} & \textbf{5.39} & \textbf{4.28}
    \\
	\bottomrule
    \toprule
    & \multicolumn{1}{c}{} & \multicolumn{2}{c}{$r=2$ yd.} &  \multicolumn{2}{c}{$r=4$ yd.} &  \multicolumn{2}{c}{$r=6$ yd.} & \multicolumn{2}{c}{$\theta=2^{\circ}$} & \multicolumn{2}{c}{$\theta=6^{\circ}$} & \multicolumn{2}{c}{$\theta=8^{\circ}$}  
    \\ 
    & \multicolumn{1}{c}{} & I-$L_{2}$ & \multicolumn{1}{c}{P-$L_{2}$}  & I-$L_{2}$ & \multicolumn{1}{c}{P-$L_{2}$} & I-$L_{2}$ & \multicolumn{1}{c}{P-$L_{2}$} & I-$L_{2}$ & \multicolumn{1}{c}{P-$L_{2}$} & I-$L_{2}$ & \multicolumn{1}{c}{P-$L_{2}$} & I-$L_{2}$ & \multicolumn{1}{c}{P-$L_{2}$}\\
    \midrule
    \multirow{9}{*}{\makecell[c]{\textbf{Football-TIP}\\ (In Yards)}}
    & Mean 
    & 8.94 & $-$ & 9.28 & $-$ & 9.69 & $-$  
    & 8.74 & $-$ & 9.12 & $-$ & 9.27 & $-$ 	 
    \\
    & Median 
    & 8.99 & $-$ & 9.39 & $-$ & 9.86 & $-$  
    & 8.80 & $-$ & 9.23 & $-$ & 9.39 & $-$ 	 
    \\
    & GMAT~\cite{zhan2018generating}
    & 4.63 & $-$ & 5.37 & $-$ & 6.44 & $-$  
    & 7.37 & $-$ & 6.98 & $-$ & 6.92 & $-$ 		
    \\
    & NAOMI~\cite{liu2019naomi}
    & 4.48 & $-$ & 4.95 & $-$ & 5.83 & $-$  
    & 7.21 & $-$ & 6.82 & $-$ & 6.70 & $-$ 	
    \\
    & Linear Fit
    & 7.18 & 7.58 & 7.01 & 6.97 & 7.08 & 9.88 
    & 8.48 & 9.77 & 7.17 & 8.40 & 7.12 & 8.04	
    \\ 
    & Vanilla LSTM~\cite{hochreiter1997long}
    & 5.26 & 6.98 & 5.96 & 5.13 & 6.47 & 6.83
    & 7.33 & 10.21 & 7.85 & 7.90 & 7.75 & 7.88
    \\  
    & Vanilla VRNN~\cite{chung2015recurrent}
    & 4.61 & 5.63 & 5.31 & 5.48 & 6.29 & 5.94
    & 6.70 & 8.27 & 6.38 & 6.98 & 6.21 & 6.72
    \\
    & INAM~\cite{qi2020imitative}
    & 4.32 & 6.01 & 4.94 & 5.52 & 6.11 & 6.34   
    & 7.19 & 8.31 & 6.85 & 7.33 & 6.62 & 7.26  	
    \\   
   \cmidrule(lr){2-14}
    & \textbf{GC-VRNN} (\textbf{Ours}) 
    & \textbf{3.95} & \textbf{4.50} & \textbf{4.68} & \textbf{4.42} & \textbf{5.19} & \textbf{4.66}
    & \textbf{5.54} & \textbf{7.58} & \textbf{5.37} & \textbf{5.88} & \textbf{5.42} & \textbf{5.71}
    \\ 
	\bottomrule
    \end{tabular}
    }
    \vspace{-1mm}
    \caption{Quantitative results on datasets Basketball-TIP (in feet) and Football-TIP (in yards). Each dataset comprises six scenarios derived from two mode settings with different radius $r$ and angel $\theta$. The best results are highlighted in bold.}
    \vspace{-3mm}
\label{tab:basketballandfootball}
\end{table*}

\begin{table}[t]
	\centering
    \scalebox{0.61}{
    \begin{tabular}{lc|cc|cc|cc}
    \toprule
    \multicolumn{1}{c}{\multirow{2}{*}{Dataset}} & \multicolumn{1}{c}{\multirow{2}{*}{Methods}} & \multicolumn{2}{c}{Easy} &  \multicolumn{2}{c}{Ordinary} &  \multicolumn{2}{c}{Hard} 
    \\
     & \multicolumn{1}{c}{}  & I-$L_{2}$ & \multicolumn{1}{c}{P-$L_{2}$} & I-$L_{2}$ & \multicolumn{1}{c}{P-$L_{2}$} & I-$L_{2}$ & \multicolumn{1}{c}{P-$L_{2}$}  
     \\
    \midrule
    \multirow{6}{*}{\makecell[c]{\textbf{Vehicle-TIP}\\ (In Pixels)}}
    & Mean 
    & 337.38 & $-$ & 282.24 & $-$ & 319.68 & $-$ 
    \\
    & Median 
    & 336.94 & $-$ & 281.58 & $-$ & 318.82 & $-$
    \\
    & Linear Fit
    & 100.53 & 139.86 & 86.69 & 97.24 & 106.43 & 113.61
    \\
    & Vanilla LSTM~\cite{hochreiter1997long}
    & 83.50 & 125.05 & 75.23 & 82.76 & 87.59 & 91.61
    \\
    & Vanilla VRNN~\cite{chung2015recurrent}
    & 88.36 & 103.21 & 70.89 & 73.54 & 95.66 & 104.34
    \\
    \cmidrule(lr){2-8}
    & \textbf{GC-VRNN} (\textbf{Ours}) 
    & \textbf{65.48} & \textbf{72.44} & \textbf{58.36} & \textbf{62.03} & \textbf{74.28} & \textbf{78.12} 
    \\
	\bottomrule
    \end{tabular}
    }
    \vspace{-1mm}
	\caption{Quantitative results (in pixels) on dataset Vehicle-TIP with three scenarios. The best results are highlighted in bold.}
    \vspace{-2mm}
	\label{tab:vehicle}
\end{table}
\begin{table}[t]
	\centering
    \scalebox{0.65}{
    \begin{tabular}{c|ccc|cc|cc|cc|cc}
    \toprule
    \multirow{2}{*}{ID} & \multicolumn{3}{c|}{GCL} & \multicolumn{2}{c}{$r=3$ ft.} &  \multicolumn{2}{c}{$r=7$ ft.} &  \multicolumn{2}{c}{$\theta=10^{\circ}$} &  \multicolumn{2}{c}{$\theta=30^{\circ}$} \\
    & ST & DL & EC
    & I-$L_{2}$ & \multicolumn{1}{c}{P-$L_{2}$} & I-$L_{2}$ & \multicolumn{1}{c}{P-$L_{2}$} & I-$L_{2}$ & \multicolumn{1}{c}{P-$L_{2}$} & I-$L_{2}$ & \multicolumn{1}{c}{P-$L_{2}$}\\
    \midrule
    1 & $\checkmark$ & &
    & 7.24 & 9.46 & 7.16 & 9.01 & 6.20 & 7.30 & 5.78 & 5.28
    \\
    2 & & $\checkmark$ &
    & 7.16 & 9.44 & 6.74 & 8.93 & 6.15 & 7.27 & 5.70 & 5.19
    \\
    3 & $\checkmark$ & $\checkmark$ &
    & 7.11 & 9.33 & 6.55 & 8.54 & 5.98 & 7.18 & 5.51 & 5.08
    \\
    4 & $\checkmark$ & & $\checkmark$ 
    & 7.10 & 9.09 & 6.30 & 7.08 & 5.90 & 6.96 & 5.41 & 4.90
    \\
    5 & & $\checkmark$ & $\checkmark$ 
    & 7.07 & 8.10 & 6.26 & 6.04 & 5.87 & 6.32 & 5.88 & 6.09
    \\
    \cmidrule(lr){1-12}
    \textbf{Ours} & $\checkmark$ & $\checkmark$ & $\checkmark$ 
    & \textbf{7.03} & \textbf{7.50} & \textbf{6.24} & \textbf{5.93} 
    & \textbf{5.86} & \textbf{6.29} & \textbf{5.39} & \textbf{4.28} \\
	\bottomrule
    \end{tabular}
    }
    \vspace{-1mm}
	\caption{Component study of three GCLs in MS-GNN. ST denotes the Static Topology GCL, DL represents the Dynamic Learnable GCL, and EC represents the Edge Conditioned GCL.}
 \vspace{-3mm}
	\label{tab:gcl}
\end{table}

\subsection{Benchmarks and Setup}
\textbf{Datasets.} 
We curate three benchmarks for the joint problem of \underline{T}rajectory \underline{I}mputation and \underline{P}rediction (TIP), and we name these three datasets with \textit{-TIP} as the suffix. More details are presented in the supplementary material.

\noindent\textbf{\textit{Basketball-TIP}}: We construct Basketball-TIP using NBA dataset~\cite{zhan2018generating}, which consists of 104,003 training sequences and 13,464 testing sequences. We design two strategies, ``circle mode'' and ``camera mode'', to replicate the realistic appearance and disappearance of players. We established six scenarios by defining three different radii $r$ (feet) or three different angles $\theta$ (degree) for better evaluation. 

\noindent\textbf{\textit{Football-TIP}}: Football-TIP is established from NFL Football Dataset\footnote{\url{https://github.com/nfl-football-ops/Big-Data-Bowl}}, which contains 10,780 training sequences and 2,492 testing sequences. Similar to Basketball-TIP, we curate six scenarios by defining three different radii $r$ (yard) and three different angles $\theta$ (degree) for evaluation.
   
\noindent\textbf{\textit{Vehicle-TIP}}: We use the Omni-MOT dataset~\cite{sun2020simultaneous} to simulate incomplete observations. Three difficulty levels are associated with the camera viewpoints: Easy, Ordinary, and Hard. With the Easy viewpoint, we have 29,239 training sequences and 6,419 testing sequences. With the Ordinary viewpoint, we have 33,831 training sequences and 7,427 testing sequences. Lastly, with the Hard viewpoint, we have 31,714 training sequences and 6,962 testing sequences.

\textbf{Evaluation Protocol.}
For Basketball-TIP and Football-TIP, we observe the first 40 frames and predict the next 10 frames. For Vehicle-TIP, we observe the first 60 frames and predict the next 30 frames. The observations are incomplete and have corresponding masks to indicate the visibility.

\textbf{Metrics.} 
To assess the imputation, we determine the average $L_2$ distance (\textbf{I-$L_{2}$}) between each agent's imputed trajectory and its corresponding ground truth over time. Similarly, for prediction evaluation, we calculate the average $L_2$ distance (\textbf{P-$L_{2}$}) between each agent's predicted trajectory and its corresponding ground truth over time. 

\textbf{Baselines.}
For both trajectory imputation and prediction, we choose the following methods: Linear Fit, Vanilla LSTM~\cite{hochreiter1997long}, 
Vanilla VRNN~\cite{chung2015recurrent}, INAM~\cite{qi2020imitative}. We also implement Mean, Median, GMAT~\cite{zhan2018generating}, and NAOMI~\cite{liu2019naomi} to compare the trajectory imputation performance.

\textbf{Implementation Details.}
\begin{figure*}[ht]
  \centering
   \includegraphics[width=0.98\linewidth]{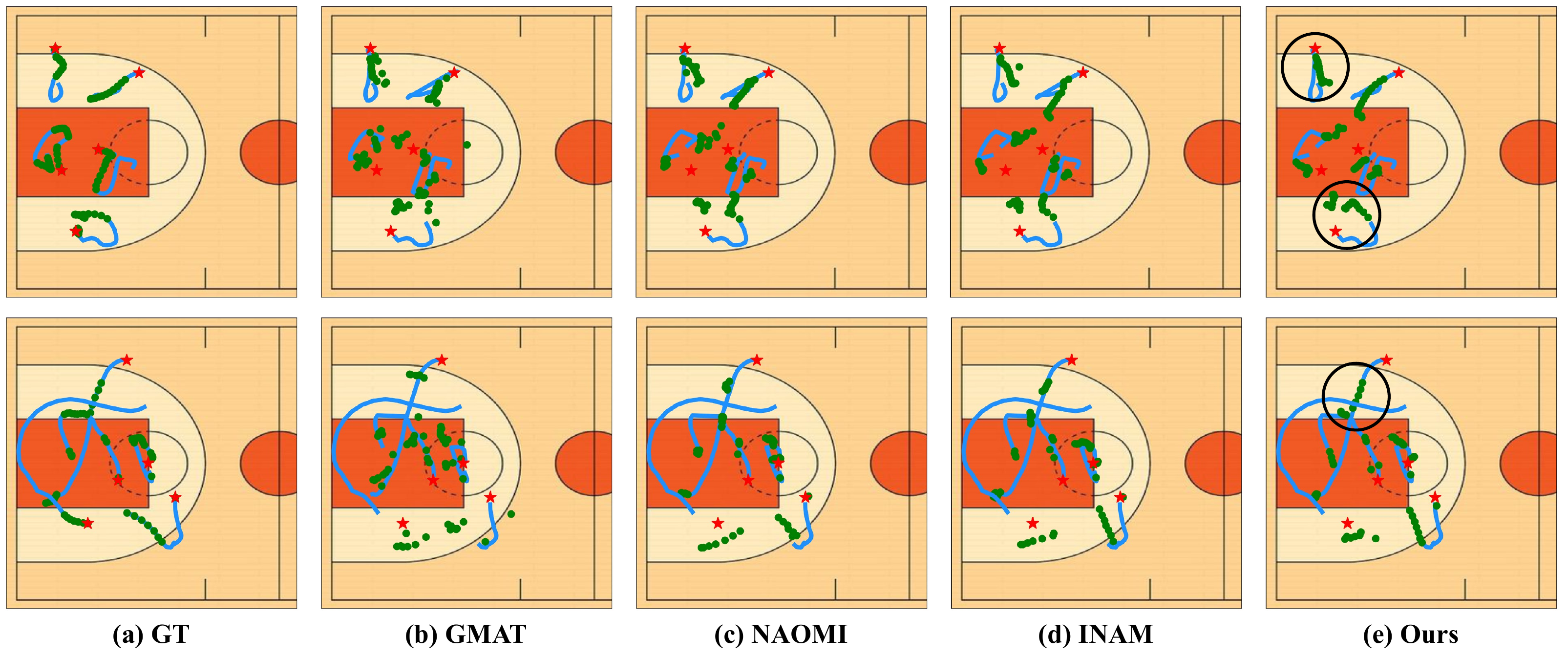}
   \vspace{-1mm}
   \caption{Visualizations of imputed results on Basketball-TIP ($\theta=30^{\circ}$). The red star denotes the starting point, the blue line represents the visible observation, and the green point represents the missing point. Note that we only plot five defenders for brevity here.}
   \vspace{-3mm}
   \label{fig:qualitative}
\end{figure*}
In MS-GNN, we stack 3 static topology GCLs of~\cref{eq:stgcl} and 1 dynamical learnable GCL of~\cref{eq:dl} for encoding both observed trajectory and future trajectory. The edge-conditioned GCL is only employed for observed trajectory and $\varphi_{ec}(\cdot)$ is a Conv2D block that includes 2 Conv2D layers with kernel size as 1, and 1 average pooling layers operating at channel level. $\varphi^{\text{pri}}(\cdot)$,  $\varphi^{\text{enc}}(\cdot)$, $\varphi^{\text{dec}}_{\cdot}(\cdot)$, and $\varphi^{\boldsymbol{z}}_{\cdot}(\cdot)$ are all implemented by MLPs. We set node feature dimension $D$ in~\cref{eq:input} and all three GCLs as 16. The RNN dimension is set as 256, and the latent variables dimension is set as 64. The experiments are conducted using PyTroch~\cite{paszke2019pytorch} on the Nvidia A100 GPU. The model is trained for 200 epochs, with a batch size of 64, utilizing the Adam optimizer~\cite{diederik2015adam} with an initial learning rate of 0.001, which decayed by 0.9 for every 20 epochs.

\subsection{Quantitative Results}
~\cref{tab:basketballandfootball} shows the quantitative results on Basketball-TIP and Football-TIP with six scenarios. In all six scenarios, our method can achieve better performance compared to other baselines, regardless of whether it pertains to the imputation or prediction task. In particular, we can observe that the improvement on the prediction task is much more significant than that on the imputation task. Unlike other baselines, valuable information is promoted to be shared via temporal flow in our framework, the imputation task can help the prediction task to a certain extent. ~\cref{sec:ablation} empirically explores the connection between these two tasks. 

~\cref{tab:vehicle} shows the quantitative results on Vehicle-TIP with three scenarios. Note that baselines~\cite{zhan2018generating, liu2019naomi, qi2020imitative} are only applicable to the sequences with a fixed number of agents, while the numbers of vehicles vary in each sequence in Vehicle-TIP. Nevertheless, our method greatly outperforms other baselines, validating that our method can effectively tackle this problem in diverse multi-agent domains.

\subsection{Ablation Study}\label{sec:ablation}
\begin{table}[t]
	\centering
    \scalebox{0.7}{
    \begin{tabular}{c|cc|cc|cc|cc}
    \toprule
    \multicolumn{1}{c|}{\multirow{2}{*}{Variants}} & \multicolumn{2}{c}{$r=3$ ft.} &  \multicolumn{2}{c}{$r=7$ ft.} &  \multicolumn{2}{c}{$\theta=10^{\circ}$} &  \multicolumn{2}{c}{$\theta=30^{\circ}$} 
    \\
    & I-$L_{2}$ & \multicolumn{1}{c}{P-$L_{2}$} & I-$L_{2}$ & \multicolumn{1}{c}{P-$L_{2}$} & I-$L_{2}$ & \multicolumn{1}{c}{P-$L_{2}$} & I-$L_{2}$ & \multicolumn{1}{c}{P-$L_{2}$}\\
    \midrule 
    w/ IMP & 7.21 & 28.74 & 6.58 & 30.06 & 5.94 & 31.84 & 5.56 & 31.57
    \\
    w/ PRE & 29.15 & 7.78 & 29.07 & 6.48 & 30.61 & 6.83 & 32.35 & 4.76
    \\
    wo/ CON & 7.11 & 16.38 & 6.40 & 13.32 & 5.96 & 15.05 & 5.48 & 13.96
    \\
    wo/ TD & 7.25 & 7.70 & 6.37 & 6.26 & 5.98 & 6.61 & 5.51 & 4.52
    \\
    \cmidrule(lr){1-9}
    \textbf{Ours}
    & \textbf{7.03} & \textbf{7.50} & \textbf{6.24} & \textbf{5.93} 
    & \textbf{5.86} & \textbf{6.29} & \textbf{5.39} & \textbf{4.28}
    \\
	\bottomrule
    \end{tabular}
    }
    \vspace{-1mm}
	\caption{Ablation study of the temporal decay module and the connection between the imputation and prediction stream.}
	\label{tab:tdconn}
 \vspace{-3mm}
\end{table}

\textbf{Three GCLs.}
~\cref{tab:gcl} shows the results of variants with different combinations of GCLs. It can be observed that each GCL contributes to the final performance of our GC-VRNN. Specifically, we can see that the edge conditioned GCL provides a greater relative improvement to the accuracy of the model than the ST and DL GCLs. This validates the effectiveness of our designed EC GCL in extracting spatial missing patterns from incomplete observations.

\textbf{Temporal Decay.} 
The TD module is designed to decipher the temporal missing patterns of incomplete observations. To study the functionality of the TD module, we remove this module for comparison, which we refer to as ``wo/ TD'' in~\cref{tab:tdconn}. It can be observed that the absence of the TD module leads to a notable decline in the performance of both tasks, confirming the effectiveness of modeling temporal missing patterns from incomplete observations. 

\textbf{Connection Between Two Streams.}
We investigate the benefits of connecting the imputation and the prediction task. We conduct the following experiments: ``w/ IMP'' means we only make imputations, and ``w/ PRE'' means we only make predictions. We also cut off the connection by introducing two different RNNs for imputation and prediction separately, which we refer to as ``wo/ CON''. It can be observed from~\cref{tab:tdconn} that considering these two tasks simultaneously can boost the performance of both tasks, especially for the prediction task. It validates the necessity of considering these two tasks in a unified framework.

\subsection{Qualitative Results}
In~\cref{fig:qualitative}, we present visual results of imputation on Basketball-TIP ($\theta=30^{\circ}$). The results demonstrate that our GC-VRNN produces more precise imputations than other baselines, thus confirming the superiority of our approach. Additional experimental outcomes, including visualizations, are available in the supplementary material.

\section{Conclusion}
Our study highlights a prevalent issue in the trajectory prediction literature, which assumes complete agent observations. We introduce a new avenue of research by jointly learning trajectory imputation and prediction. We propose a novel GC-VRNN method that uncovers spatio-temporal missing patterns and handles both tasks in a unified framework. Through experiments, we demonstrate the superiority of our designs and the benefits of simultaneously learning these tasks. To further research in this domain, we curate and benchmark three practical datasets, \textbf{\textit{Basketball-TIP}}, \textbf{\textit{Football-TIP}}, and \textbf{\textit{Vehicle-TIP}}. As far as we know, our study is the first to bridge the gap in benchmarks and techniques for this joint problem.

\newpage
\balance
{\small
\bibliographystyle{ieee_fullname}
\bibliography{main}
}

\end{document}